# Monitoring Electrostatic Adhesion Forces via Acoustic Pressure

Huacen Wang, Jiarui Zou, Zeju Zheng, and Hongqiang Wang*, *Senior Member, IEEE*

*Abstract*—Electrostatic adhesion is widely used in mobile robotics, haptics, and robotic end effectors for its adaptability to diverse substrates and low energy consumption. Force sensing is important for feedback control, interaction, and monitoring in the EA system. However, EA force monitoring often relies on bulky and expensive sensors, increasing the complexity and weight of the entire system. This paper presents an acoustic-pressure-based method to monitor EA forces without contacting the adhesion pad. When the EA pad is driven by a bipolar square-wave voltage to adhere a conductive object, periodic acoustic pulses arise from the EA system. We employed a microphone to capture these acoustic pressure signals and investigate the influence of peak pressure values. Results show that the peak value of acoustic pressure increased with the mass and contact area of the adhered object, as well as with the amplitude and frequency of the driving voltage. We applied this technique to mass estimation of various objects and simultaneous monitoring of two EA systems. Then, we integrated this technique into an EA end effector that enables monitoring the change of adhered object mass during transport. The proposed technique offers a low-cost, non-contact, and multi-object monitoring solution for EA end effectors in handling tasks.

## I. Introduction

Electrostatic adhesion (EA) is a lightweight [1], low energy consumption [2], and highly adaptable[3] adhesion technique used in crawling and climbing robotics [4]-[7], haptic devices [8]-[10], robotic handling end effectors [11]-[13]. By applying a high-amplitude and low-current voltage to parallel or coplanar interdigital electrodes embedded in the dielectric layer of EA pads, Maxwell pressures are generated to produce reliable normal and shear forces on conductive[14] or dielectric [15], [16] surfaces without mechanical clamping [17] or vacuum plumbing [18]. Its less-damaging and fast response time [19] makes EA attractive for handling delicate objects where traditional vacuum or mechanical clamps may induce damage or be impractical.

Monitoring the force of the EA-based handling system is often required, such as detecting mass change of the adhered object or identifying detachment events during static holding or the transport process. Prevailing research mostly embeds an external force or torque sensor between the EA pad and the robotic end effector [20]-[22], which increases the end effector's size, weight, complexity, and cost. Moreover, a conventional force sensor can monitor only one single EA pad [23], [24]. When multiple EA pads operate simultaneously, monitoring the whole system becomes more costly and complex.

The contributions of this work are as follows. In this work, we explored an acoustic-pressure-based method for monitoring EA forces during handling. We found that under bipolar square-wave driving, the interaction between the EA pad and the adhered object generates periodic acoustic pressure signals. By capturing these signals with a microphone positioned near the EA pad, we could monitor the mass change of the adhered object without physical contact. We also investigated the influence of key parameters on the peak acoustic pressure, concluding that it increases with the mass and contact area of the adhered object, as well as with the amplitude and frequency of the driving voltage. Furthermore, we demonstrated mass estimation of individual objects and simultaneous monitoring of two EA platforms with a single microphone. Finally, we integrated this method into an EA-based robotic end effector to monitor the weight change of the adhered object during transport.

The structure of this article is as follows. Section II outlines the basic principle of EA and the generation of acoustic pressure. Section III describes the experimental setup and explores the parameters affecting the peak acoustic pressure, along with demonstrations in three application scenarios. Section IV summarizes this article and discusses future work.

## II. Working Principle

### A. Principle of EA

The EA pad used in this work is comprised of two-phase coplanar interdigital electrodes embedded in the dielectric material, as illustrated in Fig. 1(a). When the adhered object is conductive, one phase electrode of the EA pad is driven by a high-voltage square wave (e.g., ±hundreds to thousands of volts), while the other phase is grounded, thereby generating a strong electric field between the electrodes. As the pad approaches the conductive surface of the object, image charges are induced on the substrate, producing an electrostatic adhesion force [25] that brings the two surfaces into contact, as shown in Fig. 1(b).

### B. The Generation of Acoustic Pressure

Upon bipolar square-wave excitation of the EA pad, we observed the generation of periodic acoustic pulses synchronized with each voltage polarity switch, as depicted in Fig. 1(c). The red waveform illustrates the driving voltage, while the blue trace represents the resulting acoustic pressure signal. We suppose these acoustic pressure changes are caused

*Research supported by ABC Foundation.

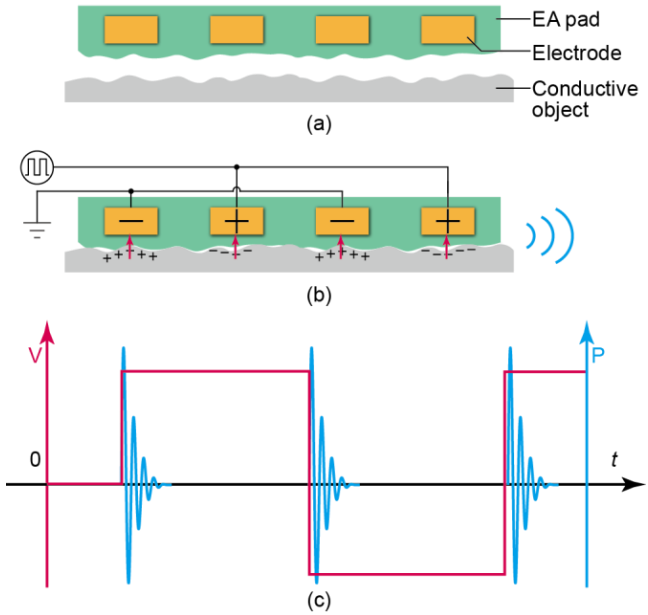

Fig. 1. The working principle of the acoustic-pressure-based method for monitoring EA forces. (a) and (b) the schematic of the principle of the EA mechanism on a conductive object. (c) The schematic of the generation of acoustic pressure.

by mechanical impacts and subsequent vibrations of the adhered object against the EA pad (Fig. 1(b)). These vibrations radiate as transient sound waves into the surrounding air [26], which are ultimately measured as acoustic pressure by a nearby microphone.

## III. EXPERIMENT RESULTS

To explore the influence parameters for the acoustic pressure generated between the EA pad and the adhered conductive object, we constructed an EA testing platform to adhere objects with varying mass and contact area, and investigated the effect of driving amplitude and frequency on the resulting acoustic pressure signals. To validate the practical applications of this work, we applied the method for mass estimation of various objects and simultaneous monitoring of two EA platforms. Then, we integrated the method into an EA-based end effector for monitoring weight changes during transport.

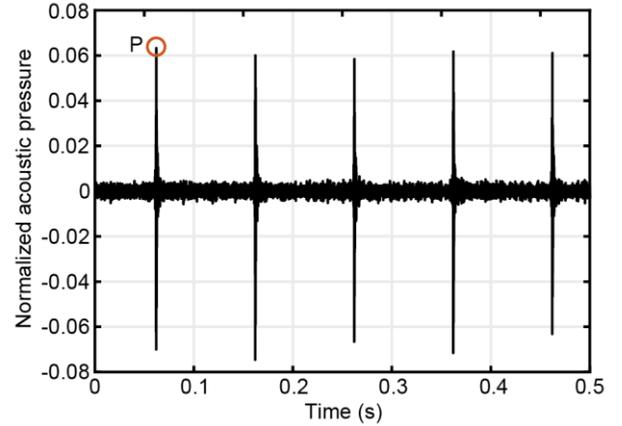

Fig. 3. The acoustic pressure signal from the EA system with a square wave supply (±600 V amplitude at 5 Hz).

### A. EA Testing Platform

The EA pad employed in this study is shown in Fig. 2(a) with dimensions of $50 \times 50 \times 1.6$ mm³. The width of each electrode and the spacing between electrodes are both 100 µm, resulting in a 200 µm pitch. The pad was fabricated using a standard PCB manufacturing process. The EA pad was adhered to a 3D-printed gantry-style frame using polyimide double-sided tape, forming an adhesion platform as depicted in Fig. 2(b), which allows objects to be adhered beneath the EA pad while driving. In the experiments, the adhered object comprised a 1.6 mm-thick aluminum plate with additional counterweights attached beneath it. A smartphone (Ace 3, OnePlus) mounted on a separate frame positioned its microphone level with the pad surface to collect the acoustic pressure signal, as shown in Fig. 2(c). MATLAB was used for data processing, filtering, and plotting. During testing, drive signals were generated by a controller board (SCB-68A, National Instruments) and amplified 1000 times by a high-voltage amplifier (Model 615-3, Trek) to drive the EA pad.

### B. Acoustic Pressure Analysis

To monitor EA adhesion forces of the testing platform, the EA pad was first driven with a square wave (±600 V amplitude at 5 Hz) to adhere an object (50 × 50 mm² contact area, 10 g mass) beneath the pad. Then, the microphone recorded raw data of acoustic pressure for several seconds. The recorded time-domain signal was first transformed to the frequency domain, and high-pass filtered at 1 kHz to suppress ambient noise. It was then converted back to the time domain to obtain the normalized acoustic pressure waveform shown in Fig. 3. The waveform exhibits five pressure peaks within 0.5 s (with a frequency of 10 Hz), which is twice the drive frequency and consistent with the supposition as depicted in Fig. 1(c). In this study, we used the peak value of acoustic pressure as the metric for EA force. The next section will analyze the influence of experimental parameters on this peak value.

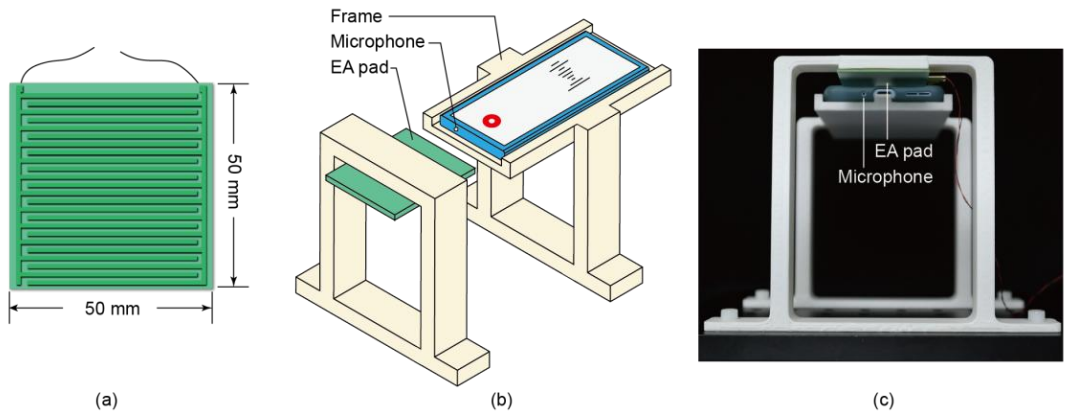

Fig. 2. The EA testing platform for adhering objects and collecting the acoustic pressure signal with a microphone. (a) The schematic of the coplanar interdigital EA pad used in this work. (b) The schematic of the EA testing platform. (c) The front view of the platform.

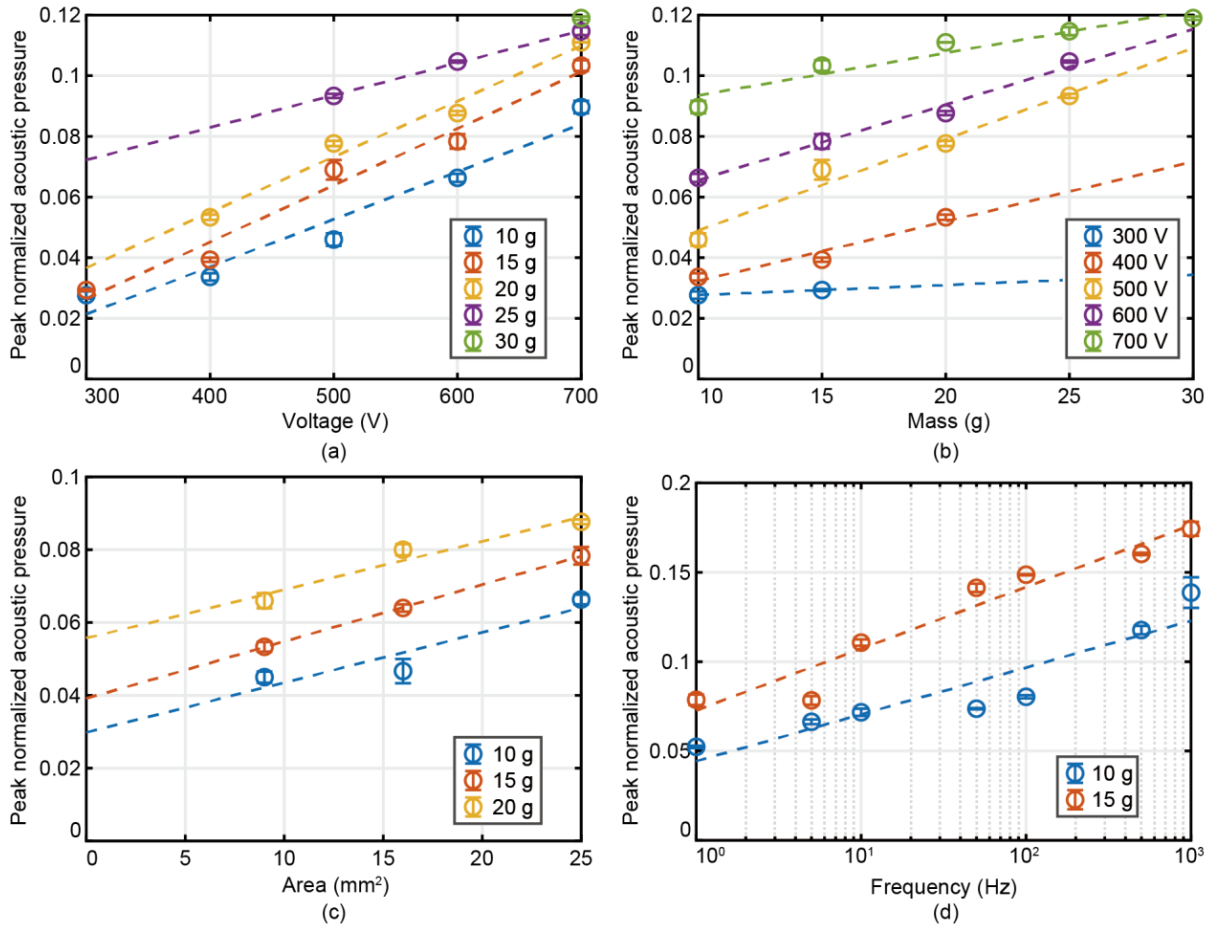

Fig. 4. The peak values of acoustic pressures under different (a) voltage amplitudes, (b) masses of adhered objects, (c) areas of object contact surfaces, and (d) driving frequencies.

### C. Results of Monitoring with Different Parameters

To explore the influence of driving voltage amplitude and adhered object mass on acoustic peak pressure, we used objects with 50 × 50 mm² contact area and maintained a fixed driving frequency of 5 Hz. The masses of the objects were 10, 15, 20, 25, and 30 g, and the amplitudes of drive voltages were set to ±300, 400, 500, 600, and 700 V. Fig. 4(a) shows the relationship between peak acoustic pressure and voltage amplitude, for a given object mass, the peak pressure increased approximately linearly with increasing voltage amplitude. Fig. 4(b) illustrates the relationship between peak acoustic pressure and object mass. For a fixed drive voltage amplitude, the peak pressure increased approximately linearly with object mass.

To investigate the influence of contact area on peak acoustic pressure, we tested objects of 10, 15, and 20 g mass with contact areas of 30 × 30, 40 × 40, and 50 × 50 mm² under a square wave (±600 V amplitude at 5 Hz). As shown in Fig. 4(c), the results show that, for a given mass, peak acoustic pressure increased with contact area. To assess the impact of drive frequency, we adhered 10 g and 15 g objects (50 × 50 mm² contact area) under a ±600 V square wave, respectively. The tested frequencies were set to 1, 5, 10, 50, 100, 500, and 1000 Hz. The experimental results are shown in Fig. 4(d), for a fixed object mass, the peak acoustic pressure grew in proportion to the logarithm of the driving frequency, and a higher object mass showed a larger growth trend.

Table 1. The comparison of the actual and estimated mass (in units of g).

|  | Nuts | Clips | Rubber duck | Tape roll | Spring |
|---|---|---|---|---|---|
| Actual mass | 1 | 8.2 | 9.5 | 12.7 | 17.6 |
| Estimated mass | 1.25 | 5.21 | 7.08 | 13.96 | 20.63 |

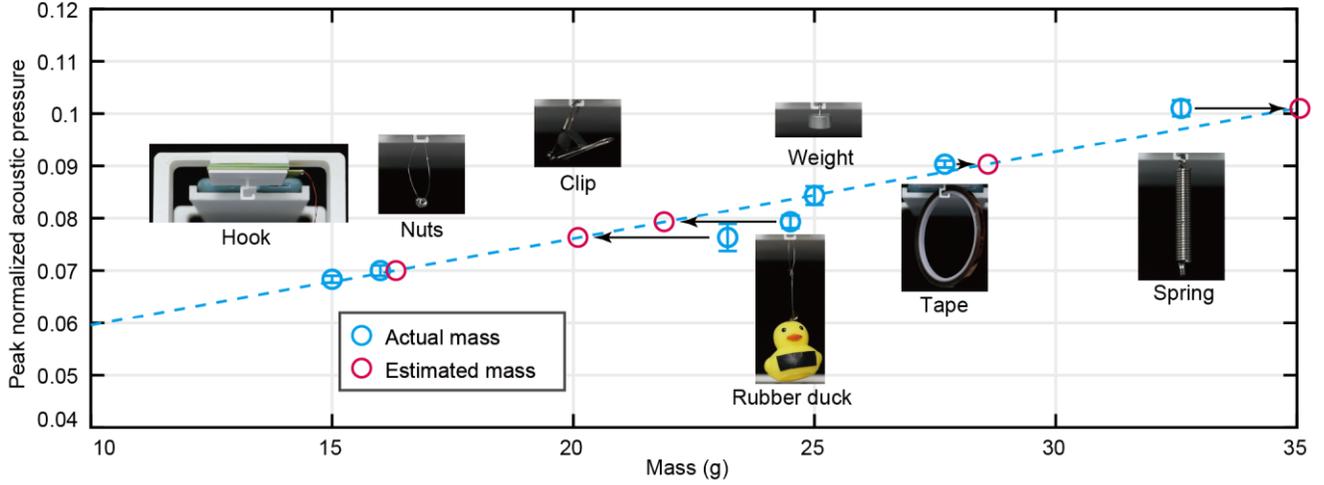

Fig. 5. The result of mass estimation of various objects based on the peak normalized acoustic pressure.

### D. Mass Estimation of Various Objects

The previous results demonstrate that, at a constant amplitude and frequency of driving voltage, peak acoustic pressure increases approximately linearly with adhered object mass. We applied this relationship to estimate object masses using a square wave (±600 V amplitude at 5 Hz) supplied to the EA pad (as shown in Movie S1).

First, an adsorbable hook (assembly with a 50 × 50 mm² aluminum plate bonded to a 3D-printed hook, with total mass about 15 g) was adhered to the EA testing platform. The recorded peak acoustic pressure was approximately 0.068, as shown in Fig. 5. A calibrated weight (mass of 10 g) was then hung on the hook, forming a total mass of 25 g and a measured peak pressure of approximately 0.084. From these two calibration points, we derived the linear relation $P = 0.0016 \cdot m + 0.044$, where $P$ is the peak value of the measured acoustic peak pressure and $m$ is the total mass of the adhered hook and suspended object. Using this relation, we first measured the peak pressures of various small objects, including nuts (with an actual mass of 1 g), clips (8.2 g), a rubber duck (9.5 g), a tape roll (12.7 g), and a spring (17.6 g) from the tests. Then, estimating the masses by the linear relation, the estimated masses were 1.25 g, 5.21 g, 7.08 g, 13.96 g, 20.63 g, respectively, indicated by red circles in Fig. 5. As listed in Table 1, the errors from comparison of actual masses and estimated masses by acoustic pressures are likely due to variations in center-of-mass of the hanging objects and resulting different vibrations.

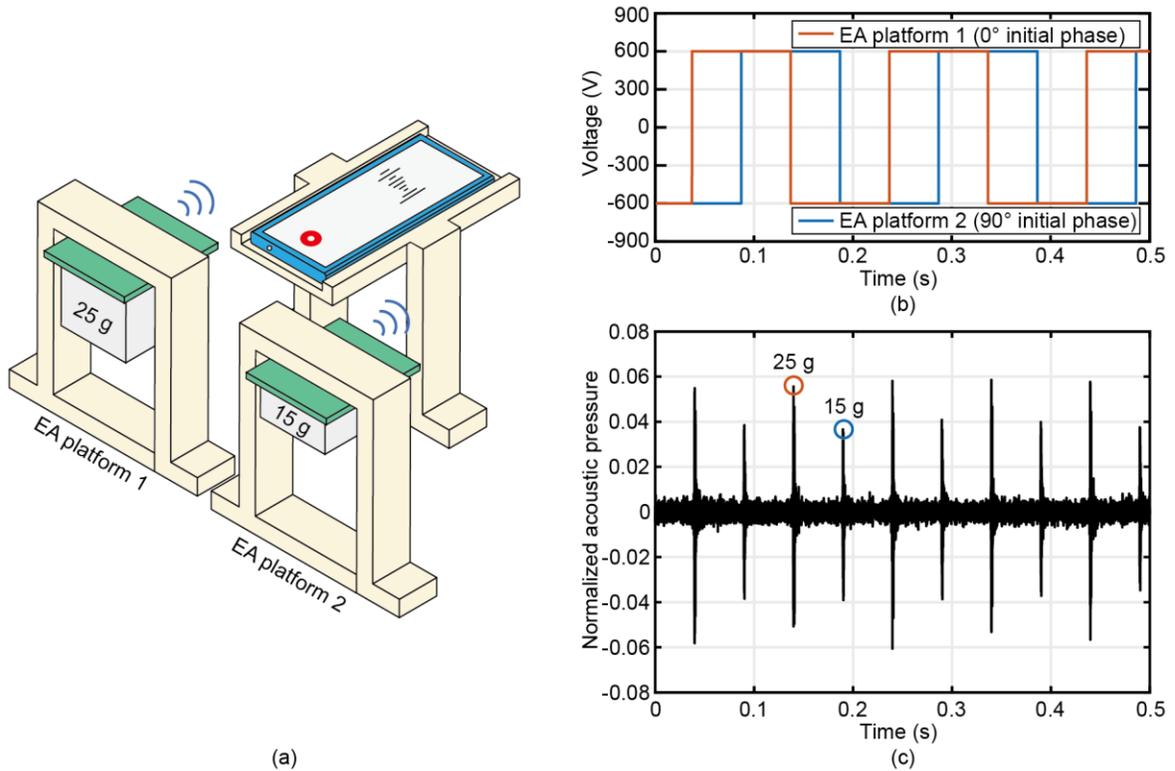

Fig. 6. Monitoring two EA platforms simultaneously based on acoustic pressure. (a) The schematic of the experiment setup. (b) The schematic of two supplied voltages with a 90° phase difference for two platforms, respectively. (c) The result of the acoustic pressure when monitoring two platforms simultaneously.

### E. Monitoring Two EA Platforms Simultaneously

Unlike conventional EA systems that require a separate force sensor for each EA pad when monitoring, our approach can use a single microphone to monitor multiple pads simultaneously. In this test, two identical EA platforms (Platform 1 and Platform 2) were employed, each adhering objects with 50 × 50 mm² contact area and masses of 25 g and 15 g, respectively (as shown in Fig. 6(a)). Both pads were driven by square waves (±600 V amplitude at 5 Hz), but with a 90° phase difference to distinguish their generation acoustic pressures (Fig. 6(b)). As shown in Fig. 6(c), the pressure signals from the two pads did not overlap in the time domain due to the initial phase difference, allowing a clear distinction of peak values of different objects. The acoustic peak pressures were approximately 0.056 for the 25 g object and 0.037 for the 15 g object. The heavier object generated a larger acoustic peak pressure, which conformed to the results from Fig. 4(b). Thus, the adhesion statuses of both platforms could be monitored simultaneously based on acoustic peak pressures.

### F. Monitoring Weight Changes during the Transport

We finally integrated the acoustic monitoring method into an end effector by attaching an EA pad and a miniature recorder (W1, Newsmy) onto a 3D-printed mount connected with the end effector. At t = 0 s, the EA pad was powered up with a square wave (±600 V amplitude at 5 Hz) and moved close to a 20 g tray with a 50 × 50 mm² conductive contact area (Fig. 7(a)). As shown in Fig. 7(b), no significant acoustic signal was recorded at this stage, even though the EA pad had been powered. Then, upon the collision between the EA pad and the tray around 1 s, large acoustic pulses were observed. During the lifting and transport process, the peak acoustic pressure remained stable, with minor fluctuations likely due to acceleration and deceleration of movements. In the middle of transport, a 10 g weight was added to the tray, which produced a noticeable increase in acoustic peak pressure. Finally, the tray was delivered to the target location, and the EA pad was powered down, returning the acoustic signal to baseline. This experiment validates that our method can monitor mass changes and detect detachment events during both adhesion and transport phases.

## IV. CONCLUSION

In summary, this work has proposed an acoustic pressure-based method for EA force monitoring, with the characteristics of low-cost, non-contact, and multi-object. The experimental results confirm that peak acoustic pressure correlates with object mass, contact area, and drive parameters (voltage amplitude and frequency). Furthermore, this article demonstrates the feasibility of this method for mass estimation, multi-object monitoring with phase-shifted excitation, and monitoring in the end effector for detection of mass changes and detachment events during transport. However, the fundamental acoustic generation mechanism and key influencing factors have only been experimentally validated, and future work should derive analytical models to better explain the phenomenon. Additional improvements, such as an advanced filtering method for enhanced noise robustness and integration of microprocessors into compact microphone modules for real-time processing, will further expand the method's robustness and applicability. This work showcases its ability and potential to monitor the EA force in handling tasks.

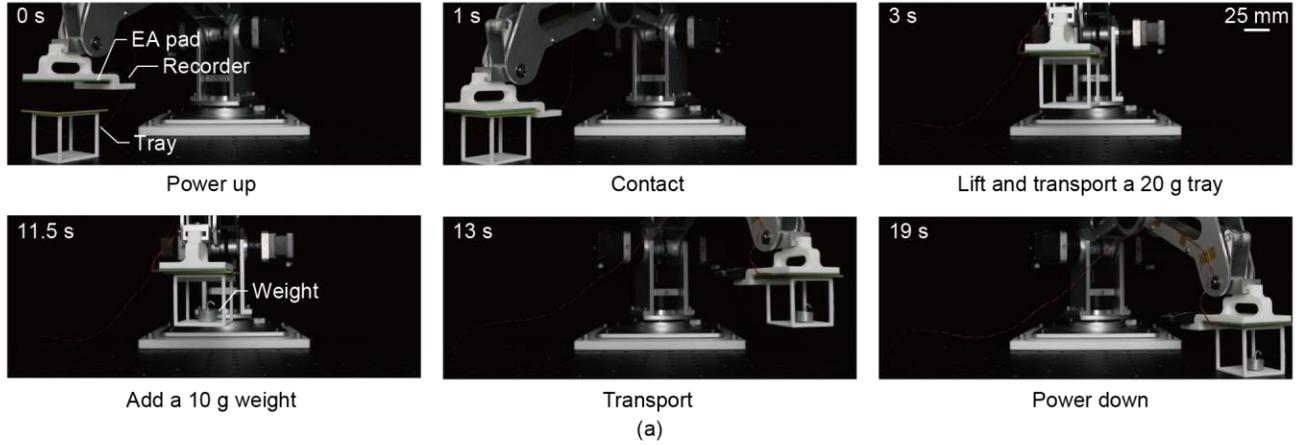
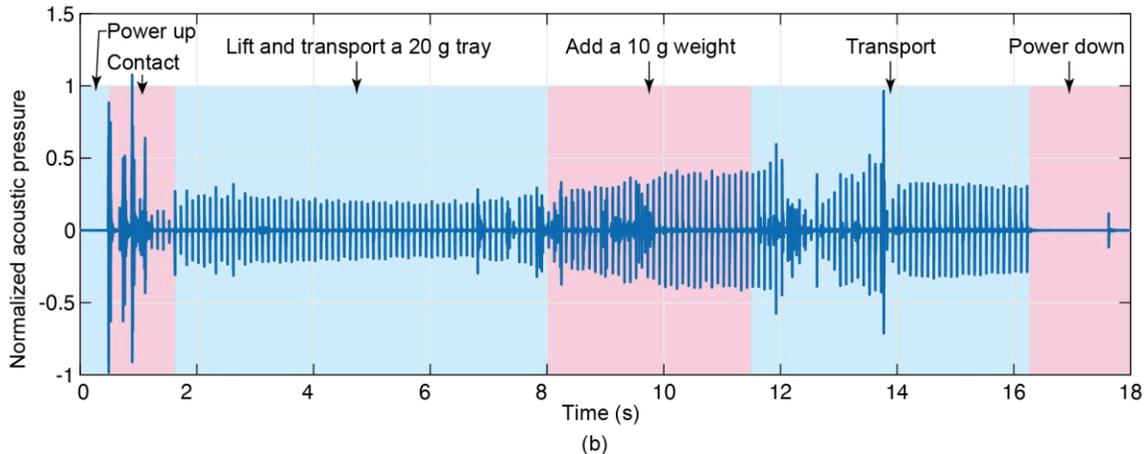

Fig. 7. Monitoring weight changes during the transport using an EA-based end effort. (a) The transport process. (c) The acoustic pressure signal during the transport process.

ACKNOWLEDGMENT